\documentclass{article}

\usepackage{amsmath}
\usepackage{amsfonts}
\usepackage{microtype}
\usepackage{graphicx}
\usepackage{subcaption}
\usepackage{multirow}
\usepackage{booktabs} 
\usepackage[colorlinks=true,bookmarks=false,citecolor=blue,urlcolor=blue]{hyperref}

\usepackage{hyperref}

\usepackage[accepted]{icml2020}

\icmltitlerunning{Increasing Input Dimensionality in Deep RL}

\begin{document}

\twocolumn[
\icmltitle{Can Increasing Input Dimensionality Improve Deep Reinforcement Learning?}




\begin{icmlauthorlist}
\icmlauthor{Kei Ota}{melco}
\icmlauthor{Tomoaki Oiki}{melco}
\icmlauthor{Devesh K. Jha}{merl}
\icmlauthor{Toshisada Mariyama}{melco}
\icmlauthor{Daniel Nikovski}{merl}
\end{icmlauthorlist}

\icmlaffiliation{melco}{Mitsubishi Electric Corporation, Kanagawa, Japan}
\icmlaffiliation{merl}{Mitsubishi Electric Research Laboratory, Cambridge, USA}

\icmlcorrespondingauthor{Kei Ota}{Ota.Kei@ds.MitsubishiElectric.co.jp}
\icmlcorrespondingauthor{Oiki Tomoaki}{Oiki.Tomoaki@dh.MitsubishiElectric.co.jp}
\icmlcorrespondingauthor{Devesh K. Jha}{jha@merl.com}
\icmlcorrespondingauthor{Toshisada Mariyama}{Mariyama.Toshisada@ab.MitsubishiElectric.co.jp}
\icmlcorrespondingauthor{Daniel Nikovski}{nikovski@merl.com}

\icmlkeywords{Machine Learning, ICML}

\vskip 0.3in
]



\printAffiliationsAndNotice{}  

\begin{abstract}
Deep reinforcement learning (RL) algorithms have recently achieved remarkable successes in various sequential decision making tasks, leveraging advances in methods for training large deep networks.
However, these methods usually require large amounts of training data, which is often a big problem for real-world applications. One natural question to ask is whether learning good representations for states and using larger networks helps in learning better policies. In this paper, we try to study if increasing input dimensionality helps improve performance and sample efficiency of model-free deep RL algorithms. To do so, we propose an online feature extractor network (OFENet) that uses neural nets to produce \textit{good} representations to be used as inputs to deep RL algorithms. Even though the high dimensionality of input is usually supposed to make learning of RL agents more difficult, we show that the RL agents in fact learn more efficiently with the high-dimensional representation than with the lower-dimensional state observations. We believe that stronger feature propagation together with larger networks (and thus larger search space) allows RL agents to learn more complex functions of states and thus improves the sample efficiency. Through numerical experiments, we show that the proposed method outperforms several other state-of-the-art algorithms in terms of both sample efficiency and performance.
Codes for the proposed method are available at \href{http://www.merl.com/research/license/OFENet}{http://www.merl.com/research/license/OFENet}.

\end{abstract}

\section{Introduction}\label{sec:intro}


Deep reinforcement learning (RL) algorithms have achieved impressive success in various difficult tasks such as computer games \citep{Mnih2013} and robotic control \citep{Gu2017}. Significant research effort in the field has led to the development of several successful RL algorithms \cite{PPO,TD3,SAC, schulman2015trust}.
Their success is partly based on the expressive power of deep neural networks that enable the algorithms to learn complex tasks from raw sensory data. Whereas neural networks have the ability to automatically acquire task-specific representations from raw sensory data, learning representations usually requires a large amount of data. This is one of the reasons that the application of RL algorithms typically needs millions of steps of data collection. This limits the applicability of RL algorithms to real-world problems, especially problems in continuous control and robotics. This has driven tremendous research in the RL community to develop sample-efficient algorithms~\cite{buckman2018sample, kurutach2018model, kalweit2017uncertainty, munos2016safe}.  

In general, state representation learning (SRL)~\cite{SRL} focuses on representation
learning where learned features are in low dimension, evolve through time, and are influenced by actions of an agent. Learning lower dimensional representation is motivated by the intuition that state of a system represents the \textit{sufficient statistic} required to predict its future and in general, \textit{sufficient statistic} for a lot of physical systems is fairly small dimensional. In the SRL framework, the raw sensory data provided by the environment where RL agents are deployed is called an \textit{ observation}, and its low-dimensional representation is called a \textit{state}. Such a state variable is expected
to have all task relevant information, and ideally only such information.
The representation is usually learned from \textit{auxiliary tasks}, that enables the state variable to contain prior knowledge of the task domain \citep{Jonschkowski2014} or the dynamics of the environment.
In contrast to auxiliary tasks, the task that the RL agents needs to learn ultimately is called the actual task in this paper.

Conventional wisdom suggests that the lower the dimensionality of the state vector, the faster and better RL algorithms will learn. This reasoning justifies various algorithms for learning compact state representations from high-dimensional observations, for example \cite{embed2control}. However, while probably correct, this reasoning likely applies to the intrinsic dimensionality of the state (the \textit{sufficient statistic}). An interesting question is whether RL problems with intrinsically low-dimensional state can benefit by intentionally increasing its dimensionality using a neural network with good feature propagation. This paper explores this question empirically, using several representative RL tasks and state-of-the-art RL algorithms.


Additionally, we borrow motivation from the fact that larger networks generally allow better solutions as they increase the search space of possible solutions. The number of units in the hidden layers of multi-layer perceptrons (MLP) is often larger than the number of inputs, in order to improve the accuracy of function approximation. The importance of the size of the hidden layers has been investigated by a number of authors.
\cite{Zhou2017} theoretically shows that MLPs with a hidden layer smaller than the input dimension are very limited in their expressive power. Also, in deep RL, neural networks often have hidden layers larger than the dimension of observations \citep{Henderson2018,fu2019diagnosing}.
Since the state representation is an intermediate variable of processing just like the hidden units, it is reasonable to expect that high-dimensional representations of state might improve the expressive power of the neural networks used in RL agents in their own right.

Based on this idea, we propose OFENet: an Online Feature Extractor Network that constructs and uses high-dimensional representations of observations and actions, which are learned in a online fashion (i.e., along with the RL policy). 
We use a neural network for OFENet to produce the representations.
It is desirable that the neural network can be optimized easily and produce meaningful high-dimensional representations. To meet these requirements, we use MLP-DenseNet as the network architecture; it is a slightly modified version of a densely connected convolutional network \citep{Densenet}. The output of MLP-DenseNet is the concatenation of all layer's outputs. This network is trained with the incentive to preserve the \textit{sufficient statistic} using an auxiliary task to predict future observations of the system. Consequently, the RL algorithm receives higher dimensional features learned by the OFENet which have \textit{good} predictive power of future observations. 

We believe that the representation trained with the auxiliary task allows our agent to learn effective, higher-dimensional representation for as input to the RL algorithm. This in turn allows the agent to learn complex policies more efficiently. We present results that demonstrate (empirically) that the representations produced by OFENet improve the performance of RL agents in non-image continuous control tasks. OFENet with several state-of-the-art RL algorithms~\cite{PPO, TD3, SAC} consistently achieves state-of-the-art performance in various tasks, without changing the original hyper-parameters of the RL algorithm. 

\section{Related work}\label{sec:related_work}
Our work is broadly motivated by~\cite{Munk2016} that proposed to use the output of a neural network layer as the input for a deep actor-critic algorithm. Our method is built on this general idea, too. However, a key difference is that the goal of representation learning in~\cite{Munk2016} is to learn compact representation from noisy observation while we propose the idea of learning \textit{good} higher-dimensional representations of state observations. For clarity of presentation, we describe the method in~\cite{Munk2016} later in detail in Section~\ref{sec:Munk}.


While the classic reinforcement learning paradigm focuses on reward maximization, streaming observation contains abundance of other possible learning targets~\cite{jaderberg2016reinforcement}. These learning tasks are known as auxiliary tasks and they generally accelerate acquisition of useful state representations. There is lot of literature on using different auxiliary tasks for different tasks~\cite{RoboticPrior,SRL,Munk2016,Zhang2018,Hoof2016, hernandez2019agent, 8793561, chen2016infogan, alvernaz2017autoencoder}. In the proposed work, we use the auxiliary task of predicting the next observation for training the OFENet with the motivation that this allows the higher dimensional outputs of the OFENet to preserve the \textit{sufficient statistic} for predicting the future observations of the system. Thus these representations can prove effective in learning meaningful policies too.

The network architecture of OFENet draws on the success of recently reported research on deep learning. In the advertisement domain, Wide \& Deep model \cite{WideDeep} and DeepFM \cite{DeepFM} have an information flow passing over deep networks, in order to utilize low-order feature interactions. Similarly, OFENet also has connections between shallow layers and the output to produce higher dimensional representations.
Contrary to ours, \cite{aravind2017towards} enriches the representational capacity using random Fourier features of the observations. Our proposed method is not orthogonal to their approach: OFENet can be combined with any types of expanded input spaces including RBF policy. That would further expand the search space to explore, and it might result in even better performance.

The most similar approach to our method has been proposed in~\cite{Zhang2018}.
Their DDR ONLINE produces higher-dimensional representations than the original observations, similar to our method. However, our approach to constructing good representations is quite different. They used a recurrent architecture of the network in order to incorporate temporal dependencies, and it contained several auxiliary tasks to increase the number of training signals. In contrast, our method uses only a single auxiliary task, and we embed additional information, such as information about the action, into the representation instead of increasing the number of training signals. Consequently, our training algorithm is much simpler than that presented in~\cite{Zhang2018}.


\section{Background}\label{sec:background}

In this section, we provide relevant background and introduce some notations that are used throughout the paper.

\subsection{Reinforcement Learning}\label{subsec:RL}
Reinforcement learning considers the setting of an RL agent interacting with an environment to learn a policy that decides the optimal action of the agent. The environment is modeled as a Markov decision process (MDP) defined by the tuple $(\mathcal{O}, \mathcal{A}, p, r)$, where 
$\mathcal{O}$ is the space of possible observation and $\mathcal{A}$ is the space of available actions. We assume that observations and actions are continuous. The unknown dynamics $p(o_{t+1}|o_t, a_t)$ represents the distribution of the next observation $o_{t+1}$ given the current observation $o_t$ and 
the current action $a_t$. The reward function $r(o_t,a_t)$ represents the reward $r_{t+1}$ obtained for $o_t$ and $a_t$. The goal of the RL agent is to acquire the policy $\pi : \mathcal{O} \rightarrow \mathcal{A}$ that maximizes the expected sum of rewards in an episode. 
Note that the word {\it state} represents the notion of the essential information in the observation in SRL, so we call the information provided by the environment {\it an observation}.

\subsection{Model learning deep deterministic policy gradient} \label{sec:Munk}
\begin{figure}[t]
	\begin{center}
		\includegraphics[clip,width=7cm]{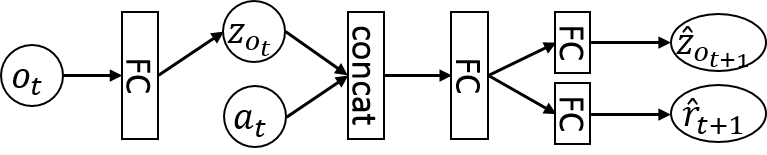}
		\caption {The model network of ML-DDPG. {\bf FC} represents a fully-connected layer, and {\bf concat} represents a concatenation of its inputs.}
		\label{img:munk}
	\end{center}
\end{figure}

\cite{Munk2016} proposed the Model Learning Deep Deterministic Policy
Gradient (ML-DDPG) algorithm to learn representations of observations.
They introduced a \textit{model network}, which is trained to construct 
the observation representation $z_{o_t}$. The representation is used as the input in the DDPG algorithm \citep{DDPG}. The model network is a three-layer network, which produces $z_{o_t}$ as an internal representation, and predicts the next observation representation $\hat{z}_{o_{t+1}}$ and the reward $\hat{r}_{t+1}$ from the observation $o_{t}$ (Figure\ref{img:munk}). 

The model network is trained by minimizing the following loss function
\begin{equation*}
L_m = ||z_{o_{t+1}} - \hat{z}_{o_{t+1}}||^2 + \lambda_m ||r_{t+1} - \hat{r}_{t+1}||^2,
\end{equation*}
where $\lambda_m$ represents the trade-off between predicting the reward and the next observation representation. The minimization is done with samples collected before the agent starts learning, and then the parameters of the model network are fixed during learning.

While we construct high-dimensional representations with OFENet, in the experiments of ML-DDPG, the dimension of the observation representation $z_{o_t}$ was not greater than one third of the dimension of the observation $o_{t}$.

\section{Online feature extractor network}\label{sec:OFENet}

In this section, we describe our proposed method for learning higher-dimensional state representations for training RL agents. In the standard reinforcement learning setting, an RL agent interacts with the environment over a number of discrete time steps. At any time $t$, the agent receives an observation  $o_t$ along with a reward $r_t$ and emits an action $a_t$. During training, standard RL agents receive observation-action pair as input to learn the optimal policy. We propose the Online Feature Extractor Network (OFENet), which constructs high-dimensional representations for observation-action pair, which is then used by the RL algorithm as input to learn the policy (instead of the raw observation and action pair). The observation used by the components is replaced by the observation representation $z_{o_t}$, and the observation-action pair is also replaced by the observation-action representation $z_{o_t,a_t}$.

\begin{figure}[t]
	\centering
	\includegraphics[width=\columnwidth]{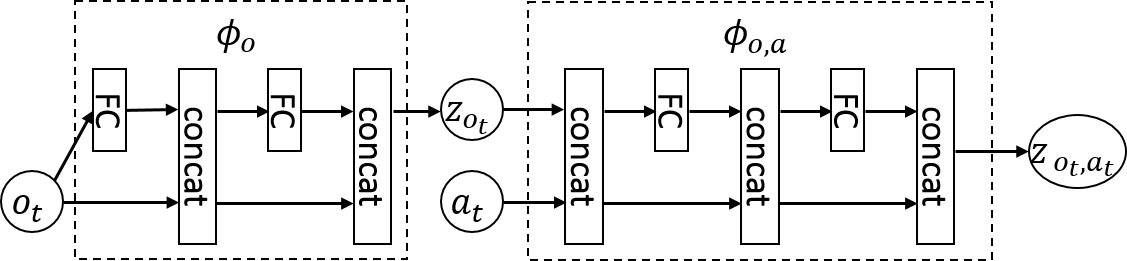}
	\caption {An example of the online feature extractor. 
		{\bf FC} represents a fully connected layer with an activation function, 
		and {\bf concat} represents a concatenation of the inputs.
	}
	\label{dataflow}
\end{figure}

OFENet learns the mappings $z_{o_t}=\phi_o(o_t)$ and $z_{o_t,a_t}=\phi_{o,a}(o_t,a_t)$, which have parameters $\theta_{\phi_o}, \theta_{\phi_{o,a}}$ as depicted in Figure~\ref{dataflow}.
To learn the mappings, we use an auxiliary task whose goal is to predict the next observation from the current observation and action.
The learning of the auxiliary task is done concurrently with the learning of the actual task (and thus we call our proposed network as Online Feature Extraction Network, OFENet).


In the following, we describe in detail the auxiliary task, the neural network architecture for the mappings $\phi_{o}, \phi_{o,a}$, and how to select hyper-parameters for OFENet.

\subsection{Auxiliary task}  \label{sec:auxtasks}

In this section, we incorporate \textit{auxiliary tasks} in order to learn effective higher-dimensional representations for state and action to be used by the RL agent. It is common knowledge that incorporating auxiliary tasks into the reinforcement learning framework can promote faster training, more robust learning and ultimately higher performance for RL agents~\citep{mujoco, jaderberg2016reinforcement}.

We introduce the module $f_{pred}$, which receives the observation-action representation $z_{o_t,a_t}$ as input to predict the next observation $o_{t+1}$. 
The module $f_{pred}$ is represented as a linear combination of the representation $z_{o_t, a_t}$, which has parameters $\theta_{pred}$. 

Thus, effectively, along with learning the actual RL objective, we optimize the parameter set $\theta_{aux} = \{\theta_{pred}, \theta_{\phi_o}, \theta_{\phi_{o,a}}\}$ to minimize the auxiliary task loss defined as:
\begin{equation}\label{eqn:aux_loss}
L_{aux} = \mathbb{E}_{(o_t, a_t) \sim p,\pi} [||f_{pred}(z_{o_t,a_t}) - o_{t+1}||^2]
\end{equation}


The auxiliary task defined by Equation~\eqref{eqn:aux_loss} is used by the OFENet to learn higher-dimensional representations. This loss function incentivizes higher-dimensional representations that preserve the dynamics information of the system (or loosely preserve the sufficient statistic to predict future observations of the system). The expectation is that the RL agent can now learn much more complex policies using these higher-dimensional features as this effectively increases the search space for the parameters of the policies.

The transitions $(o_t, a_t, o_{t+1})$ required for learning are sampled as mini-batches $\{o_{t,\mathcal{B}}, a_{t,\mathcal{B}}, o_{t+1, \mathcal{B}}\}$ from the experience replay buffer $\mathcal{B}$, which stores the past transitions that the RL agent has received. Algorithm \ref{alg1} outlines this procedure.

\begin{algorithm}[t]
	\caption{Training of OFENet}         
	\label{alg1}
	Initialize parameters $\theta_{aux} = \{\theta_{pred}, \theta_{\phi_o}, \theta_{\phi_{o,a}}\}$
	\newline
	Initialize experience replay buffer $\mathcal{B}$ 
	
	\begin{algorithmic}[1]
		\FOR{each environment step}
		\STATE $a_t \sim \pi(a_t | o_t)$
		\STATE $o_{t+1} \sim p(o_{t+1}|o_t, a_t)$
		\STATE $\mathcal{B} \Leftarrow \mathcal{B} \cup \{(o_t, a_t, o_{t+1}, r_{t+1})\}$
		\STATE sampling mini-batch $\{o_{t,\mathcal{B}}, a_{t,\mathcal{B}}, o_{t+1, \mathcal{B}}\}$ from $\mathcal{B}$
		\STATE $\theta_{aux} \Leftarrow \theta_{aux} - \lambda_{\theta_{aux}} \nabla_{\theta_{aux}} L_{aux}$
		\STATE resampling mini-batch $\{o_{t,\mathcal{B}}, a_{t,\mathcal{B}}, o_{t+1, \mathcal{B}}\}$ from $\mathcal{B}$
		\STATE $z_{o} \Leftarrow \phi_o(o_{t,\mathcal{B}})$
		\STATE $z_{o,a} \Leftarrow \phi_{o,a}(z_o, a_{t,\mathcal{B}})$
		\STATE Update the agent (e.g., SAC) parameters with the representations $z_{o}, z_{o,a}$
		\ENDFOR
	\end{algorithmic}
\end{algorithm}

\subsection{Network architecture} \label{sec:architecture}
A neural network is used to represent the mappings $\phi_{o}, \phi_{o,a}$ in OFENet. As it is known that deeper networks have advantages with respect to optimization ability and expressiveness, we employ them in our network architecture.
In addition to this, we also leverage the fact that observations often have intuitively useful information in non-image RL tasks. For example, when position and velocity of a robot are present in an observation, it is advantageous to include them in the representation when solving reaching tasks. Moreover, because the linear combination of position and velocity can approximate the position for the next time step, outputs of shallow layers are also expected to be physically meaningful.

To combine the advantages of deep layers and shallow layers, we use MLP-DenseNet, which is a slightly modified version of DenseNet \citep{Densenet}, as the network architecture of OFENet. Each layer of MLP-DenseNet has an output $y$ which is the concatenation of the input $x$ and the product of a weight matrix $W_1$ and $x$ defined as:
\begin{equation*}
y = [x, \sigma(W_1 x)]
\end{equation*}
where $[x_1, x_2]$ means concatenation, $\sigma$ is the activation function, and the biases are omitted to simplify notation. Since each layer's output is contained in the next layer's output, the raw input and the outputs of shallow layers are naturally contained in the final output.

The mappings $\phi_{o}, \phi_{o,a}$ are represented with an MLP-DenseNet. The mapping $\phi_{o}$ receives the observation $o_t$ as input, and the mapping $\phi_{o,a}$ receives the concatenation of the observation representation $z_{o_t}$ and the action $a_t$ as its input. Figure \ref{dataflow} shows an example of these mappings in the proposed OFENet.

RL algorithms take the learned representations $z_{o_t}$ and $z_{o_t,a_t}$ as input, and compute the optimal policy by optimizing the regular objective of maximizing the expected reward. It is important to note that these representations, however, are learned simultaneously with the RL algorithms. This might lead to change in distribution of the inputs to the RL algorithm. The RL algorithm, therefore, needs to adapt to the possible change of distribution in those base layers constructed by MLP-DenseNet. To alleviate this potential problem, we normalize the output of the base layer by using Batch Normalization \cite{batch_norm} to suppress changes in input distributions.

\subsection{Hyperparameter selection} \label{sec:select}

Effective training of agents with OFENet requires selection of size of the hidden layers and the type of activation functions. In~\cite{Henderson2018}, authors show that the size of the hidden layers and the type of activation function can greatly affect the performance of RL algorithms, and the combination that achieves the best performance depends strongly on the environment and the algorithm.
Thus, ideally, we would like to choose the architecture of OFENet by measuring the performance on the actual RL task, but this would be very inefficient.
Therefore, we use the performance on the auxiliary task as an indicator for selecting the architecture for each task.

In order to measure performance on the auxiliary task, first we collect transitions using a random action policy for the agent.
The transitions are randomly split into a training set and a test set.
Then, we train each architecture on the training set, and measure the average loss $L_{aux}$ on the test set over five different random seeds. On the actual task, we use the architecture which achieves the minimum on the average auxiliary loss. We call this average loss the {\it auxiliary score}. Since we can reuse the transitions for each OFENet training, and do not have to train RL agents, this procedure is sample-efficient and doesn't incur much computational cost either.

We use an experience replay buffer to simulate the learning with the RL agent, where OFENet samples a mini-batch up to the $N$th data item of the training set at $N$th step.

The psuedo-code for the proposed method is presented in Algorithm~\ref{alg1}. It is noted that we do not tune the hyperparameters of the RL algorithm in order to show that agents can learn effective policies by using the representations learned by the proposed method during the learning process. This allows more flexibility in training of RL agents across a wide range of tasks. 

\section{Experiments}\label{sec:experiments}
In this section, we try to answer the following questions with our experiments to describe the performance of OFENet.

\begin{itemize}
    \item What is a good architecture that learns effective state and state-action representations for training better RL agents?
    \item Can OFENet learn more sample efficient and better performant polices when compared to some of the state-of-the-art techniques? 
    \item What leads to the performance gain obtained by OFENet?
    \item How does the dimensionality of OFENet representation affects performance?
\end{itemize}

In the rest of this section, we present experiments designed to answer the above questions. All these experiments are done in MuJoCo simulation environment.
\subsection{Architecture comparison} \label{exp:comparison}
We compare the auxiliary score defined in section \ref{sec:select} with the performance on the actual task for various network architectures on Walker2d-v2 task, where the dimensions of observations and actions are respectively $17$ and $6$.

First, we define the {\it actual score}, which is the metric of performance on the actual task. In this paper,

\begin{itemize}
	\item {\it return} represents a cumulative reward over an episode;
	\item {\it step score} represents the average return over 10 episodes with the RL agent at each step;
	\item {\it actual score} represents the average of the step scores over latest 100,000 steps, where the step score is measured every 5,000 steps.
\end{itemize}

We measure the auxiliary score and the actual score for each network architecture.
In this section, each architecture is characterized by a connectivity of architecture, number of layers, and an activation function. We compare three connectivity architectures: MLP-DenseNet defined in section \ref{sec:architecture}, standard MLP, and MLP-Resnet, which is a slightly modified version of ResNet \citep{Resnet}.
MLP-ResNet has skip connections similar to the original one, and its layers have the output $y$ defined as:
\begin{equation}
y = \sigma (W_2 \sigma(W_1 x) + x)
\end{equation}
where $W_1, W_2$ are weight matrices, $x$ is the input, and $\sigma$ is the activation function. The biases are omitted to simplify notation.

Each architecture has multiple options for the combination of a layer number and a hidden layer size. In this experiment, $\phi_o, \phi_{o, a}$ have the same layer number for each architecture. MLP has 1, 2, 3, or 4 layers for $\phi_o$. MLP-ResNet and MLP-DenseNet have 2, 4, 6, or 8 layers for $\phi_o$.

To find the most efficient architecture over same feature size, the dimensions of $z_{o_t}, z_{o_t,a_t}$ are respectively fixed to $137$ and $263$. This means that the dimensionality increments of $z_{o_t}, z_{o_t,a_t}$ from their inputs are $120$. While the numbers of hidden units in $\phi_o, \phi_{o, a}$ are respectively $137$ and $263$ in MLP and MLP-Resnet, the number of hidden units in MLP-DenseNet depends on the number of layers. All the layers in MLP-DenseNet have the same number of hidden units. For example, when $\phi_o$ has 4 layers, the number of hidden units is $120/4=30$.

Additionally, we compare the following activation functions: ReLU, tanh, Leaky ReLU, Swish \citep{swish} and SELU \citep{selu}. In total, we compare $3$ connectivity architectures, $4$ layer-size combinations, and $5$ activation functions, resulting in a total of $60$ network architectures.

To measure the auxiliary score, we collect 100K transitions as a training set and 20K transitions as a test set, using a random policy. Each architecture is trained for 100K steps. To measure the actual score, each architecture is trained with the SAC agent for 500K steps with Algorithm \ref{alg1}. The SAC agent is trained with the hyper-parameters described in \cite{SAC}, where the networks have two hidden layers which have 256 units. All the networks are trained with mini-batches of size 256 and Adam optimizer, with a learning rate $3\cdot10^{-4}$.

\begin{figure}[t]
	\begin{center}
		\includegraphics[clip,width=7cm]{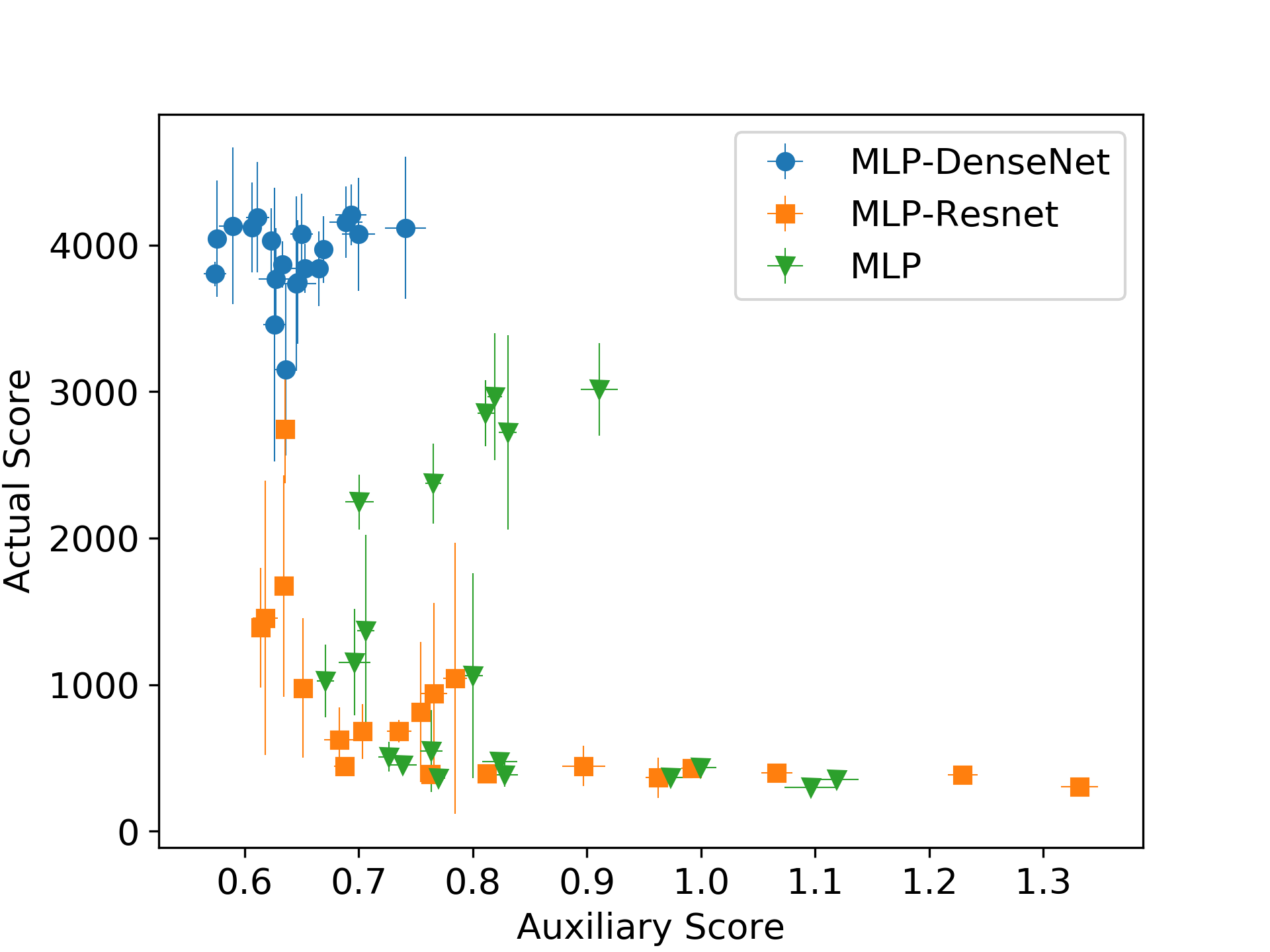}
		\caption {The actual scores and the auxiliary scores of various architectures in Walker2d-v2. The error bars represent the standard deviation of 5 trials.}
		\label{img:archselect}
	\end{center}
\end{figure}

Figure \ref{img:archselect} shows the actual score and the auxiliary score of each network architecture in Walker2d-v2. The results show that DenseNet consistently achieves higher actual scores than other connectivity architectures. With respect to the auxiliary score, DenseNet also achieves better performance than others in many cases.

Overall, we can find a weak trend that the smaller the auxiliary scores, the better the actual scores. 
Therefore, in the following experiment, we select the network architecture that has the smallest value of the auxiliary score among the 20 DenseNet architectures for the actual task for each environment.

\subsection{Comparative evaluation}

\begin{figure*}[t]
	\begin{minipage}[]{0.32\linewidth}
		\centering
		\includegraphics[width=6.1cm]{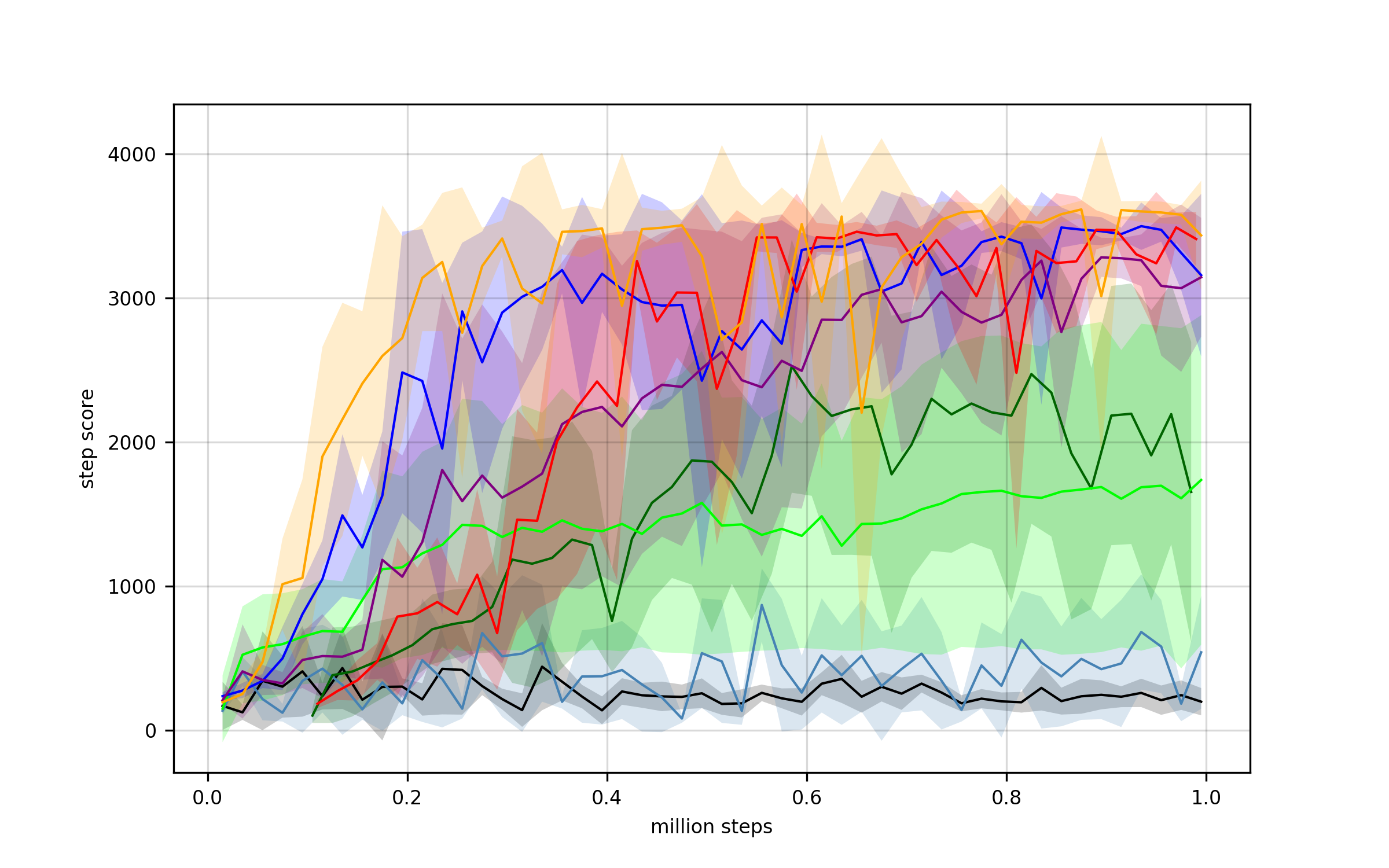}
		\subcaption{Hopper-v2}\label{fig:hopper}
	\end{minipage}
	\begin{minipage}[]{0.32\linewidth}
		\centering
		\includegraphics[width=6.1cm]{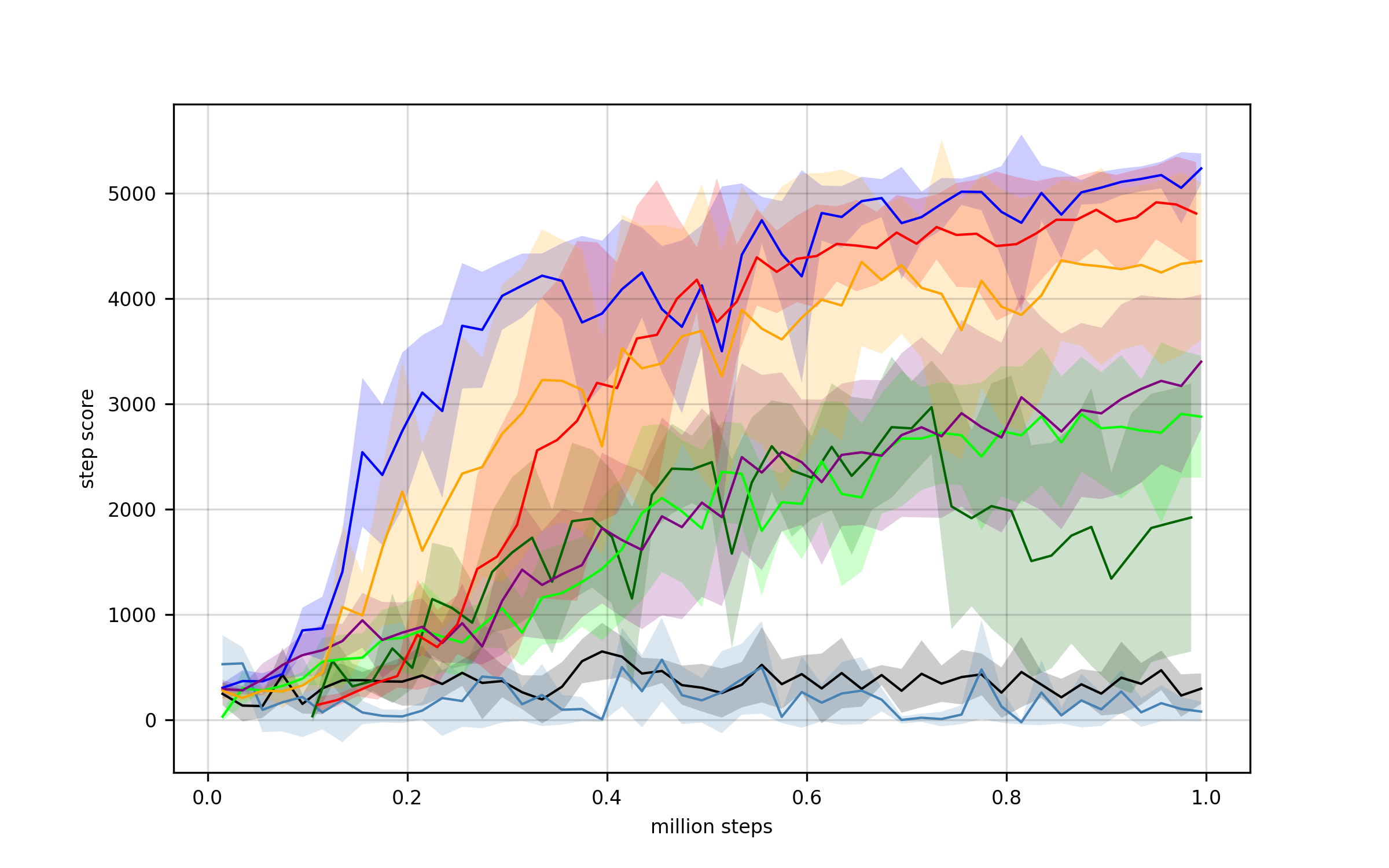}
		\subcaption{Walker2d-v2}\label{fig:walker2d}
	\end{minipage}
	\begin{minipage}[]{0.32\linewidth}
		\centering
		\includegraphics[width=6.1cm]{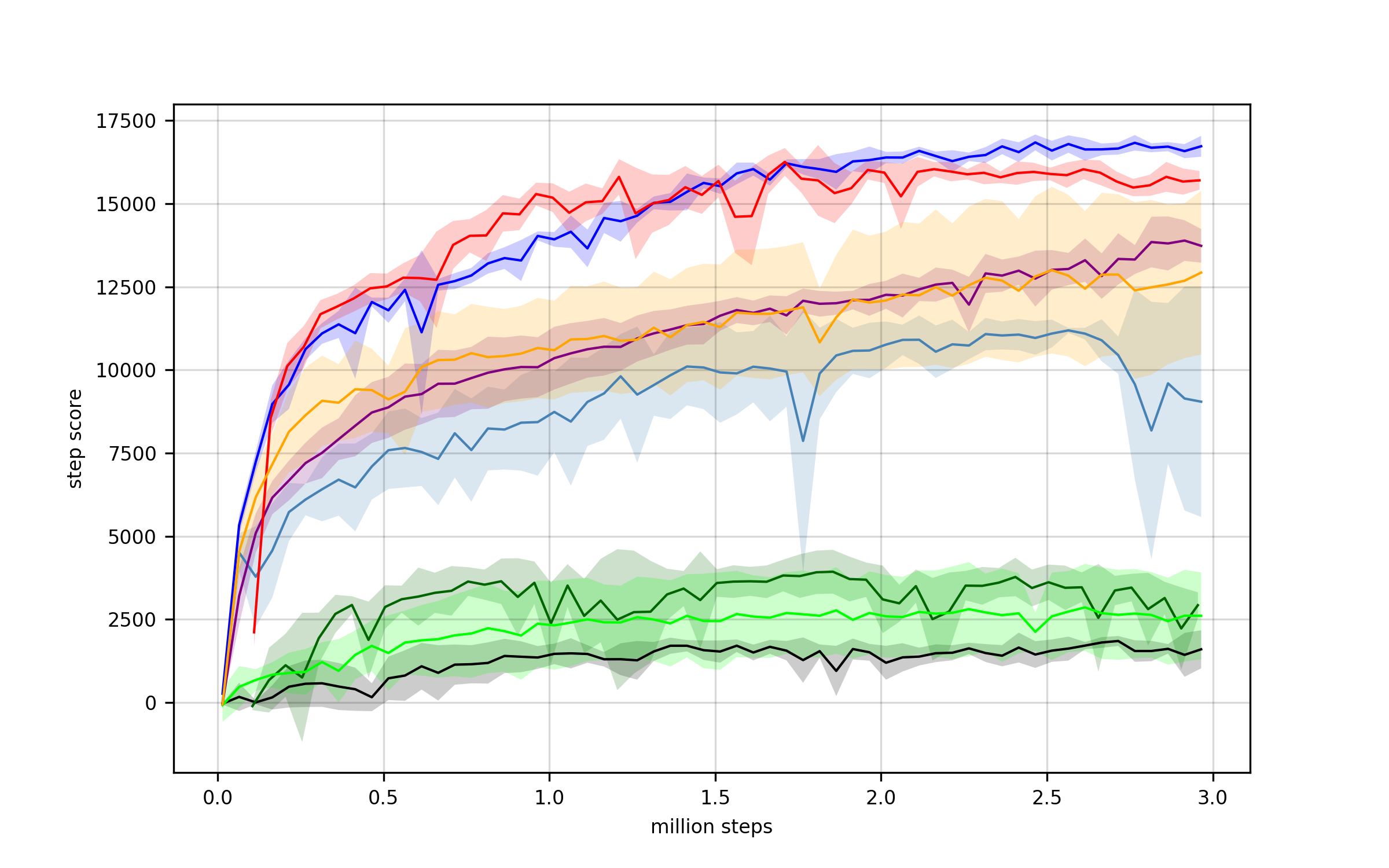}
		\subcaption{HalfCheetah-v2}\label{fig:halfcheetah}
	\end{minipage} \\
	\begin{minipage}[]{0.32\linewidth}
		\centering
		\includegraphics[width=6.1cm]{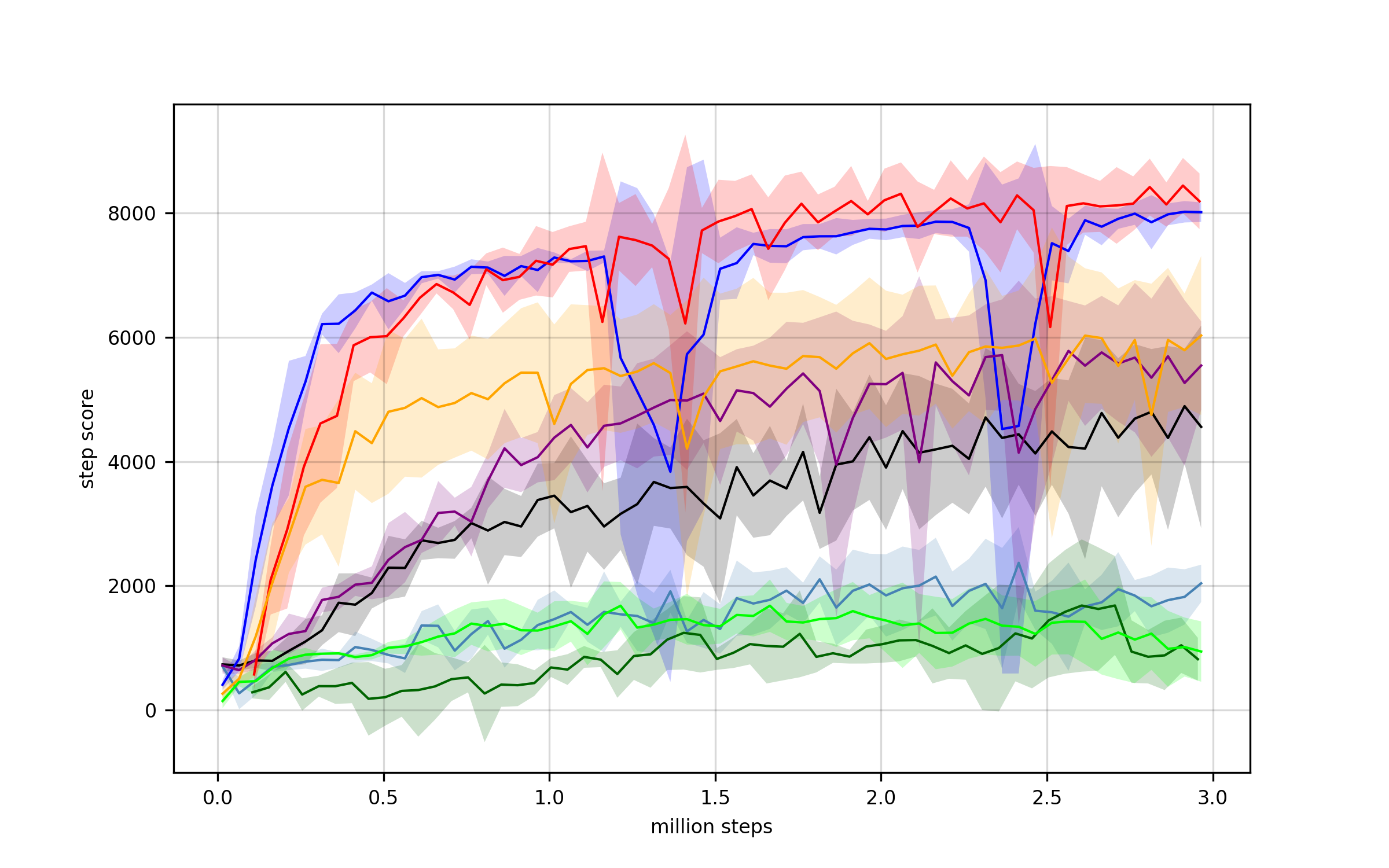}
		\subcaption{Ant-v2}\label{fig:Ant}
	\end{minipage}
	\begin{minipage}[]{0.32\linewidth}
		\centering
		\includegraphics[width=6.1cm]{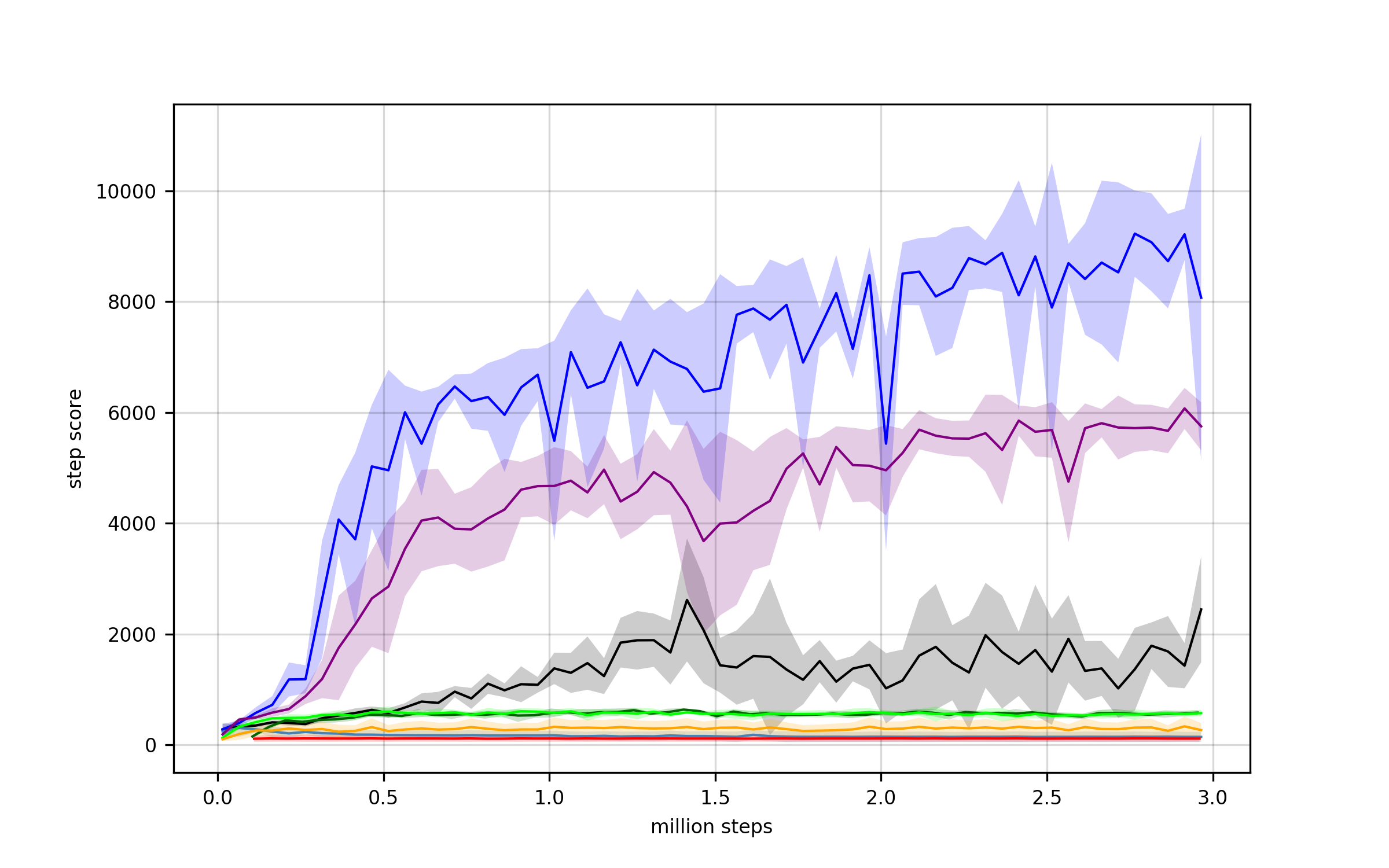}
		\subcaption{Humanoid-v2}\label{fig:humanoid}
	\end{minipage}	
	\begin{minipage}[]{0.32\linewidth}
	\centering
	\includegraphics[width=3cm]{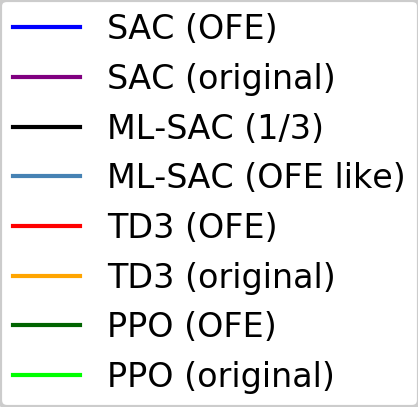}
    \end{minipage}	

	\caption{Training curves on OpenAI Gym tasks. The solid lines represent average returns over five instances with different random seeds. The shaded region represents the standard deviation of the five instances. OFE outperforms original algorithm on both on-policy (PPO) and off-policy methods (SAC and PPO).}
	\label{fig:actual}
\end{figure*}

\begin{table*}[t]
    \caption{The highest average returns over five different seeds for each environment. The bold number indicates the best score among each algorithm (SAC, TD3, and PPO). OFE outperforms original algorithm in most environments.}
    \vskip 0.15in
    \begin{center}
    \begin{small}
    \begin{sc}
    \begin{tabular}{l|cccc|cc|cc} 
        \toprule
         & \multicolumn{4}{c}{SAC} & \multicolumn{2}{c}{TD3} & \multicolumn{2}{c}{PPO} \\
         & OFE & \multirow{2}{*}{Original} & ML-SAC & ML-SAC & OFE & \multirow{2}{*}{Original} & OFE & \multirow{2}{*}{Original} \\
         & (ours) & & (1/3) & (OFE like) & (ours) & & (ours) \\
         \midrule
         Hopper-v2 & {\bf 3511.6} & 3316.6 & 750.5 & 868.7 & 3488.3 & {\bf 3613.0} & {\bf 2525.6} & 1753.5 \\
         Walker2d-v2 & {\bf 5237.0} & 3401.5 & 667.4 & 627.4 & {\bf 4915.1} & 4515.6 & {\bf 3072.1} & 3016.7 \\
         HalfCheetah-v2 & {\bf 16964.1} & 14116.1 & 1956.9 & 11345.5 & {\bf 16259.5} & 13319.9 & {\bf 3981.8} & 2860.4\\
         Ant-v2 & {\bf 8086.2} & 5953.1 & 4950.9 & 2368.3 & {\bf 8472.4} & 6148.6 & {\bf 1782.3} & 1678.9\\
         Humanoid-v2 & {\bf 9560.5} & 6092.6 & 3458.2 & 331.7 & 120.6 & {\bf 345.2} & {\bf 670.3} & 652.4 \\
         \bottomrule
    \end{tabular}
    \end{sc}
    \end{small}
    \end{center}
    \vskip -0.1in
    \label{tab:actual}
\end{table*}

To evaluate OFENet, we measure the performance of SAC, twin delayed deep deterministic policy gradient algorithm (TD3) \citep{TD3} for off-policy RL algorithms, and proximal policy optimization (PPO) \citep{PPO} for on-policy RL algorithm with OFENet representations and raw observations, on continuous control tasks in the MuJoCo environment.
The dimensionality increments of $z_{o_t}, z_{o_t,a_t}$ from their inputs are $240$ in all experiments, 
and we select the best network architecture for each task as described in section \ref{sec:select}.
The network architectures and optimizer, hyperparameters of SAC, TD3, PPO are the same as used in their original papers \cite{SAC,TD3,PPO} even we combine them with OFENet.
The mini-batch size in \cite{TD3}, however, is different from original paper. We use mini-batches of size $256$ instead of $100$, similarly to SAC.

Moreover, we measure the performance of SAC with the representations which are produced by a model network of ML-DDPG, which we call \textit{ML-SAC}. The hidden layer size of ML-SAC is $100$, its activation function is ReLU, and $\lambda_m$ in section \ref{sec:Munk} is $10$. We train the model network with an Adam optimizer, with a learning rate $1\cdot10^{-3}$.
We set the dimension of the observation representation to one third of that of the observation according to \cite{Munk2016}. In addition to this, we measure the performance of ML-SAC with the observation representation which has the same dimension as OFENet.
Whereas in \cite{Munk2016} the model network was trained with samples collected before the learning of the agent, we train the network with samples collected by the learning agent, such as OFENet.

In order to eliminate dependency on the initial parameters of the policy, we use a random policy to store transitions to the experiment replay buffer for the first 10K time steps for SAC, and 100K time steps for TD3 and PPO as described in \cite{TD3}. We also pretrain OFENet to stabilize input to each RL algorithm with these randomly trained samples.
Note that as described in section \ref{sec:auxtasks}, OFENet predicts the future observation to learn the high-dimensional representations. In Ant-v2 and Humanoid-v2, the observation contains the external forces to bodies, which are difficult to predict because of their discontinuity and sparsity. Thus, OFENet does not predict these external forces.

Figure \ref{fig:actual} shows the learning curves of the methods, and Table~\ref{tab:actual} shows the highest average returns over five different seeds. 
SAC (OFE), i.e. SAC with OFENet representations, outperforms SAC (raw), i.e. SAC with raw observations. Especially in Walker2d-v2, Ant-v2, and Humanoid-v2, the sample efficiency and final performance of SAC (OFE) outperform significantly those of the original SAC. Since TD3 (OFE) and PPO (OFE) also outperform original algorithm, it can be concluded that OFENet is an effective method for improving deep RL algorithms on various benchmark tasks.

 
ML-SAC (1/3), i.e. ML-SAC with low dimensional representation performed poorly on all tasks. Since ML-DDPG is supposed to find compact representations from noisy observations, the model network probably could not find a compact representation from the non-redundant observations in the tasks.
ML-SAC (OFE like), i.e., ML-SAC with the high dimensional representations, also performed poorly. In addition to this, extracting representation with MLP got much worse actual scores than MLP-DenseNet in section \ref{sec:select}. These show that constructing high dimensional representations is not a trivial task, and OFENet resolves this difficulty with MLP-DenseNet.

\subsection{Ablation study}
    \newcommand{\ofe}{{\it full }}
    \newcommand{\noaux}{{\it no-aux }}
    \newcommand{\nobn}{{\it no-bn }}
    \newcommand{\raw}{{\it original }}
    \newcommand{\sameparams}{{\it same-params }}
    \newcommand{\freezed}{{\it freeze-ofe }}
    Our hypothesis is that we can extract effective features from an auxiliary task in environments with a low-dimensional observation vectors. Furthermore, we would like to verify that just increasing the dimensionality of the state representation will not help the agent to learn better policies and that, in fact, generating \textit{effective} higher dimensional representations using the OFENet is required to get better performance. To verify this, we conducted an ablation study to show what components does the improvement of OFENet comes from.
    
\begin{figure}[htbp]
    	\begin{center}
    		\includegraphics[clip,width=8.5cm]{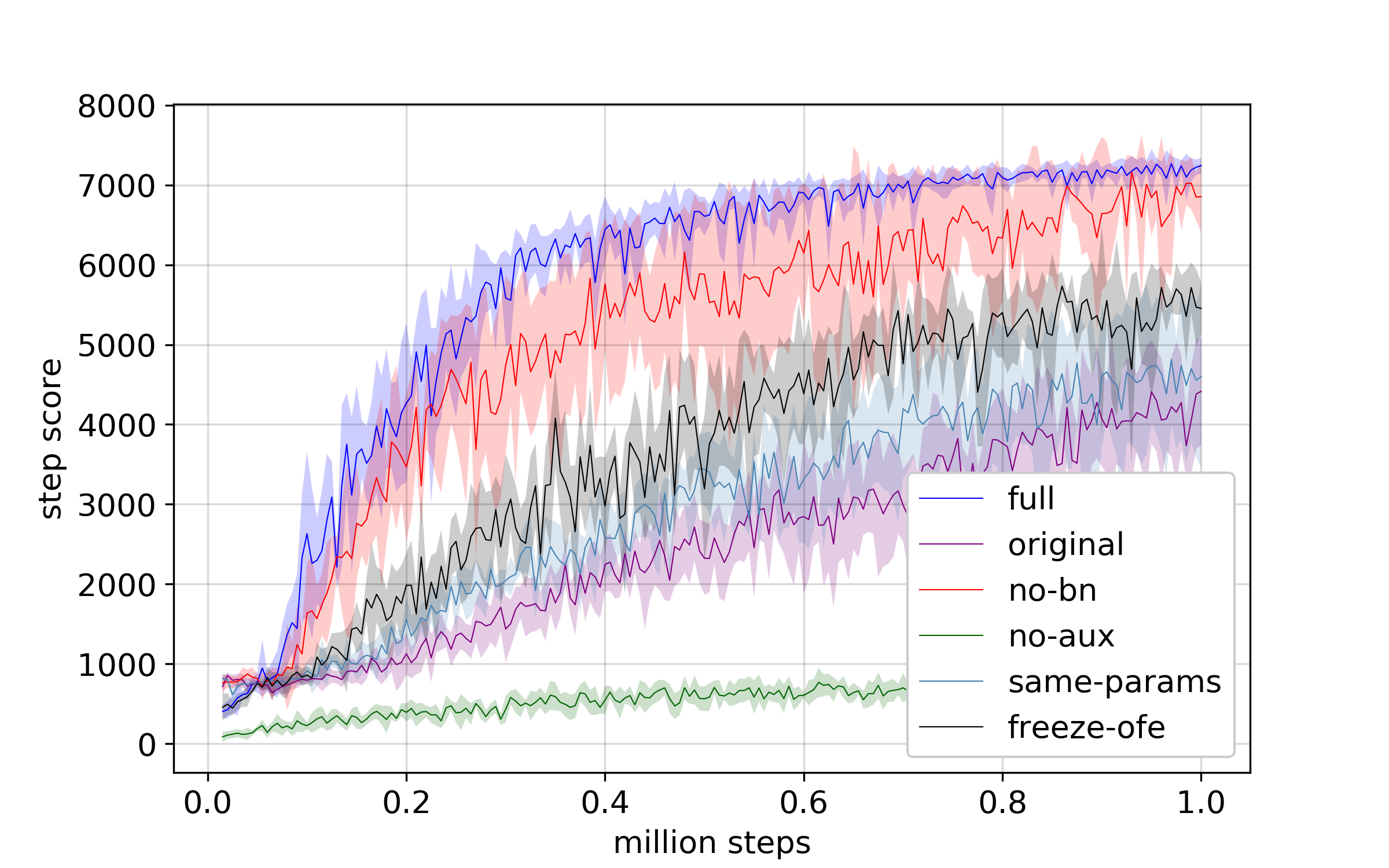}
    		\caption{Training curves of the derived methods of SAC on Ant-v2. This shows that just increasing the input dimensionality of the input representation doesn't help in learning better policies. The higher-dimensional representations need to learned with the auxiliary task proposed in the paper.}
    		\label{fig:ablation_study}
    	\end{center}
    \end{figure}

    Figure~\ref{fig:ablation_study} shows the ablation study over SAC with Ant-v2 environment.
    \ofe and \raw are the same plots of SAC (OFE) and SAC (original) from Figure.~\ref{fig:actual}.

    \nobn removes Batch Normalization from OFENet. The bigger standard deviation of \nobn indicates that adding Batch Normalization stabilizes the full learning process. Since the OFENet is learned on-line, the distribution of the input to the RL algorithms changes during training. Batch Normalization effectively works to suppress this covariate shift during training and thus the learning curve of \ofe is more stable than \nobn.

    \noaux removes auxiliary task and train both OFENet and RL algorithms with actual task objective of reward maximization. In other words, they have to be learned by only scalar reward signal. The much lower scores of \noaux shows that learning the complex OFENet structure from just the reinforcement signal is difficult, and using the auxiliary task for learning good high-dimensional features enables better learning of control policy.

    \sameparams increases the number of units of original SAC to $(401, 401)$, instead of $(256, 256)$ as suggested in~\cite{SAC} so that it has the same number of parameters with our algorithm.
    The performance does increase compared to the original unit size, but its still not as good as the full algorithm in terms of both sample efficiency and performance.
    This shows that just increasing the number of parameter does not help improve performance, but the auxiliary task helps with efficient exploration in the bigger parameter space.  

    As done in~\cite{Munk2016}, \freezed trains OFENet only before training of RL agent with randomly collected transitions as discussed in Section~\ref{exp:comparison}, and does not simultaneously train OFENet along with RL policy (i.e., skip line $5$ and $6$ in Algorithm.\ref{alg1}).
    Since the accuracy of predicting future observation becomes worse when an RL agent explores unseen observation space, freezing the OFENet trained with only randomly collected data cannot produce good representations.

    

\subsection{Effect of Dimensionality of Representation}
\begin{figure}[htbp]
	\includegraphics[clip,width=8.5cm]{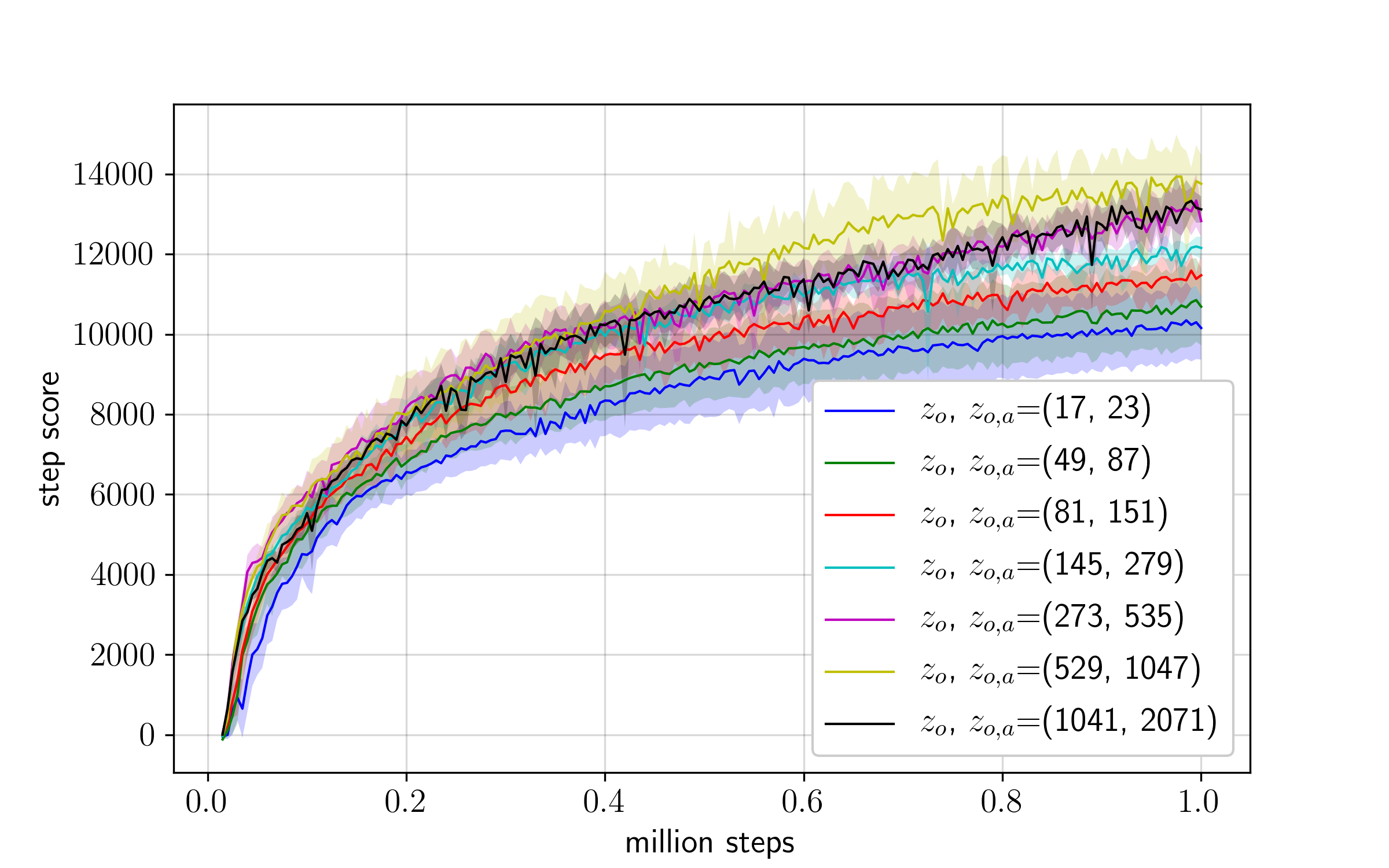}		
	\caption {Comparison of various dimensional representations on HalfCheetah-v2. This shows that increasing the size of the OFENet representation generally helps to improve the policy performance.}
	\label{img:dimensionality}
\end{figure}

In this section, we try to test whether increasing the dimension of the OFENet representation could lead to monotonic improvements in the performance of the RL agent. Figure~\ref{img:dimensionality} shows the improvement in the performance of an SAC agent on the HalfCheetah-v2 environment when we increase the dimension of the OFENet representation by increasing the numbers of hidden units in an 8-layer OFENet from $4$ to $128$. The step score of the RL agent generally increases with the increase of the dimensionality of representation, until a threshold is reached. This shows that we need sufficient output dimensionality to get the benefit of increasing the dimensionality of state representations using any feature extraction network.  


\section{Conclusion}\label{sec:conclusions}
So, can increasing input dimensionality improve deep reinforcement learning? Recent success of deep learning has allowed us to design RL agents that can learn very sophisticated policies for solving very complex tasks. It is common belief that allowing smaller state representation helps in learning complex RL polices. In this paper, we wanted to challenge this hypothesis with the motivation that larger feature representations for state can allow bigger solution space and thus, can find better policies for RL agents. To demonstrate this, we presented an Online Feature Extractor Network (OFENet), a neural network that provides much higher-dimensional representation for originally low-dimensional input to accelerate the learning of RL agents.
Contrary to popular belief we provide evidence suggesting that representations that are much higher-dimensional than the original observations can significantly accelerate learning of RL agents. However, it is important to note that the high-dimensional representations should be learned so as to retain some knowledge of the task or the system. In the current paper, it was learned using the auxiliary task of predicting the next observation.
Our experimental evaluation demonstrated that OFENet representations can achieve state-of-the-art performance on various benchmark tasks involving difficult continuous control problems using both on-policy and off-policy algorithms. Our results suggest that RL tasks, where the observation is low-dimensional, can benefit from state representation learning. Additionally, the feature learning by OFENet doesn't require tuning the hyper-parameters of the underlying RL algorithm. This allows flexible design of RL agents where the feature learning is separated from policy learning. 

In the future, we would like to study the proposed network for high-dimensional inputs to RL agents (e.g., images). We would also like to make the proposed method work with several other on-policy techniques. Additionally, we would like to study better techniques to find the optimal architectures for the OFENet to add more flexibility in the learning process. 

\nocite{langley00}

\bibliography{cite}
\bibliographystyle{icml2020}


\clearpage
\onecolumn
\appendix

\section{Appendix}

\subsection{Network architecture}
    Table.~\ref{tab:ofenet_architecture} shows the network architecture of the OFENet for each environment.
    As described in Section.~\ref{exp:comparison}, the selected network architecture is the one that receives the smallest value of the auxiliary score among 20 DenseNet architectures: number of layers are selected from $\{2, 4, 6, 8\}$, and the activation function is selected from \{ReLU, tanh, Leaky ReLU, Swish, SELU\}.

    \begin{table}[htbp]
        \caption{The network architectures of OFENet for each MuJoCo task.}
        \vskip 0.15in
        \begin{center}
        \begin{small}
        \begin{sc}
        \begin{tabular}{l|cc} 
            \toprule
             & Number of layers & Activation function \\
             \midrule
             Hopper-v2 & 6 & Swish \\
             Walker2d-v2 & 6 & Swish \\
             HalfCheetah-v2 & 8 & Swish \\
             Ant-v2 & 6 & Swish \\
             Humanoid-v2 & 8 & Swish \\
             \bottomrule
        \end{tabular}
        \end{sc}
        \end{small}
        \end{center}
        \vskip -0.1in
        \label{tab:ofenet_architecture}
    \end{table}

    The network architecture of the RL algorithms \{ SAC, TD3, PPO \} are the same with their original papers~\cite{SAC,TD3,PPO}.

\subsection{Hyper-parameters}
    Like the network architecture, the hyper-parameters of the RL algorithms are also the same with their original papers, except that the TD3 uses the batch size $256$ instead of $100$ similarly to SAC.

    Table.~\ref{tab:param_sac_ofe} shows the number of parameters for SAC (OFE) used in the experiments.
    The number of parameters of \sameparams in Figure~\ref{fig:ablation_study} matches the sum of the parameters for OFENet and the number of parameters of SAC.
    Note that the OFENet uses MLP-DenseNet architecture, and the output units of OFENet in Table.~\ref{tab:param_sac_ofe} ignores the units of previous layer.

    \begin{table}[htbp]
        \caption{The parameter size of SAC (OFE) for Ant-v2.}
        \vskip 0.15in
        \begin{center}
        \begin{small}
        \begin{sc}
        \begin{tabular}{ll|cccc} 
            \toprule
             & & Input units & Output units & Parameters \\
             \midrule
             \multirow{7}{*}{OFENet: $z_o$}
             & $1$st layer & $111$ & $40$ & $4640$ \\
             & $2$nd layer & $151$ & $40$ & $6240$ \\
             & $3$rd layer & $191$ & $40$ & $7840$ \\
             & $4$th layer & $231$ & $40$ & $9440$ \\
             & $5$th layer & $271$ & $40$ & $11040$ \\
             & $6$th layer & $311$ & $40$ & $12640$ \\
             & Total & & & $\mathbf{51840}$\\
             \midrule
             \multirow{4}{*}{SAC}
             & $1$st layer & $351$ & $256$ & $90112$ \\
             & $2$nd layer & $256$ & $256$ & $65792$ \\
             & Output layer & $256$ & $8$ & $2056$ \\
             & Total & & & $\mathbf{157960}$ \\
             \midrule
             SAC (OFE) & Total & & & $\mathbf{209800}$ \\
             \bottomrule
        \end{tabular}
        \end{sc}
        \end{small}
        \end{center}
        \vskip -0.1in
        \label{tab:param_sac_ofe}
    \end{table}






\end{document}